\documentclass[lettersize,journal]{IEEEtran}
\usepackage{amsmath,amsfonts}
\usepackage{algorithmic}
\usepackage{algorithm}
\usepackage{array}
\usepackage[caption=false,font=normalsize,labelfont=sf,textfont=sf]{subfig}
\usepackage{textcomp}
\usepackage{stfloats}
\usepackage{url}
\usepackage{verbatim}
\usepackage{graphicx}
\usepackage{cite}
\usepackage{times}
\usepackage{epsfig}
\usepackage{amssymb}
\usepackage{multicol}
\usepackage{multirow}
\usepackage{amstext}
\usepackage{booktabs}
\usepackage{float}
\usepackage{bm}
\usepackage{color}
\usepackage{fancyhdr}
\usepackage{balance}
\usepackage{overpic}
\usepackage{bbm}
\begin{document}

\title{Learning Feature Recovery Transformer for Occluded Person Re-identification}

\author{Boqiang~Xu,
        Lingxiao~He,~\IEEEmembership{Member,~IEEE,}
        Jian~Liang,
        and~Zhenan~Sun,~\IEEEmembership{Senior Member,~IEEE,}
        
\thanks{Boqiang Xu, Jian Liang and Zhenan Sun are with the Center for Research on Intelligent Perception and Computing, National Laboratory of Pattern Recognition, Institute of
Automation, Chinese Academy of Sciences, Beijing 100190, China and also with University of Chinese Academy of Sciences, Beijing 100190 (email: boqiang.xu@cripac.ia.ac.cn; liangjian92@gmail.com; znsun@nlpr.ia.ac.cn).}
\thanks{Lingxiao He is with the AI Research of JD, Beijing 100020, China (email: helingxiao3@jd.com)}}

\markboth{Journal of \LaTeX\ Class Files,~Vol.~14, No.~8, August~2021}%
{Shell \MakeLowercase{\textit{et al.}}: A Sample Article Using IEEEtran.cls for IEEE Journals}

\IEEEpubid{0000--0000/00\$00.00~\copyright~2021 IEEE}

\maketitle

\begin{abstract}
One major issue that challenges person re-identification (Re-ID) is the ubiquitous occlusion over the captured persons. There are two main challenges for the occluded person Re-ID problem, \boldsymbol{$i.e.,$} the interference of noise during feature matching and the loss of pedestrian information brought by the occlusions.
In this paper, we propose a new approach called Feature Recovery Transformer (FRT) to address the two challenges simultaneously, which mainly consists of visibility graph matching and feature recovery transformer.
To reduce the interference of the noise during feature matching, we mainly focus on visible regions that appear in both images and develop a visibility graph to calculate the similarity.
In terms of the second challenge, based on the developed graph similarity, for each query image, we propose a recovery transformer that exploits the feature sets of its $k$-nearest neighbors in the gallery to recover the complete features.
Extensive experiments across different person Re-ID datasets, including occluded, partial and holistic datasets, demonstrate the effectiveness of FRT.
Specifically, FRT significantly outperforms state-of-the-art results by at least 6.2\% Rank-1 accuracy and 7.2\% mAP scores on the challenging Occluded-Duke dataset. The code is available at \textcolor{blue}{https://github.com/xbq1994/Feature-Recovery-Transformer.}
\end{abstract}


\section{Introduction}
\begin{figure}[t]
 \centering
 \includegraphics[width=\linewidth]{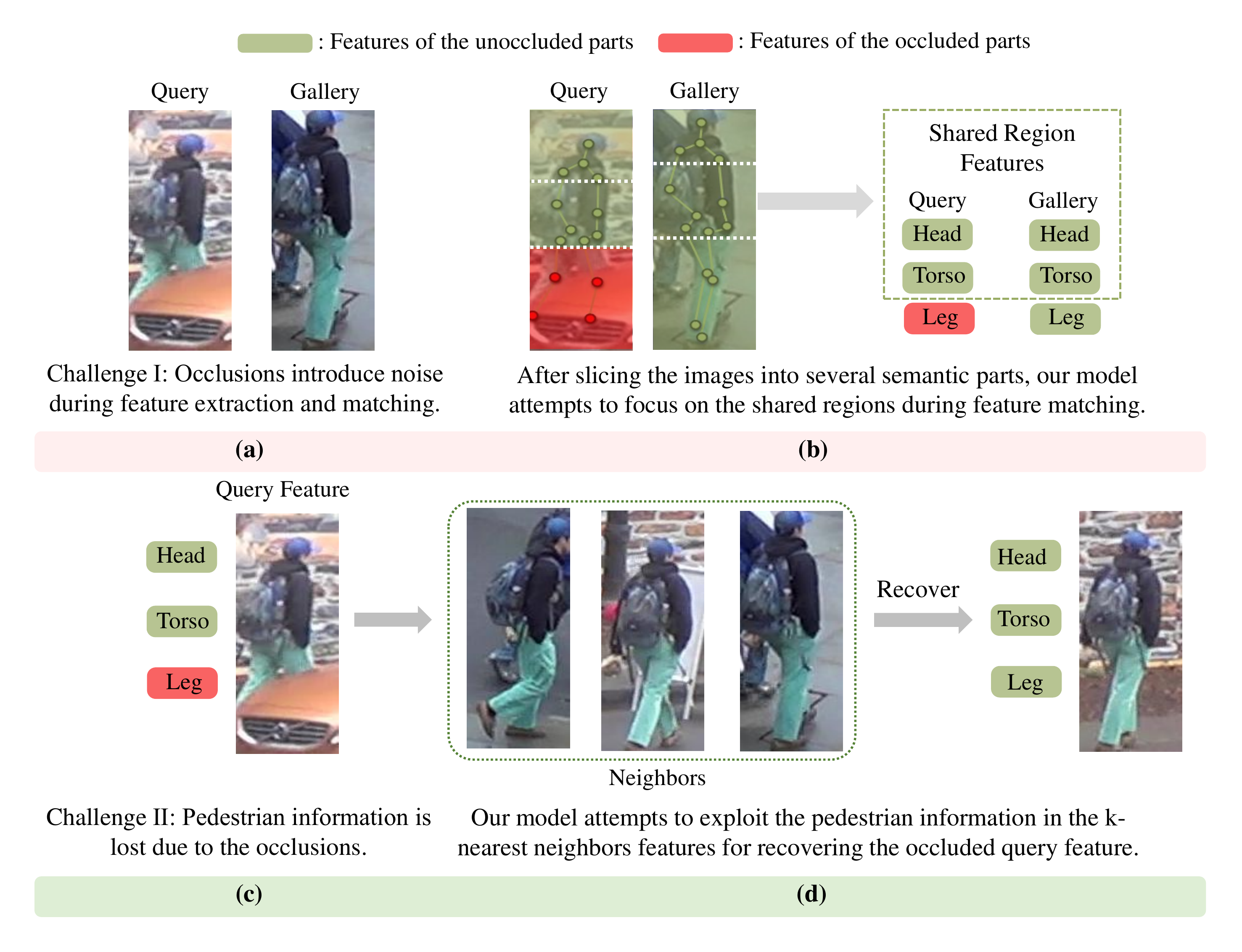}
 \caption{Two challenges in the occluded Re-ID problem and our solutions. The figure is only for illustration, our method is processed in the feature level.}
 \label{fig:motivation}
\end{figure}

\IEEEPARstart{P}{erson} re-identification (Re-ID) \cite{li2019learning,ristani2018features,xu2020black} aims to retrieve the same person from overlapping cameras, which is widely used in security, video surveillance and smart city. 
Recently, considerable Re-ID methods have been proposed in this field \cite{Jin_2020_CVPR,sun2018beyond,zhao2017deeply,Zhou_2020_CVPR}. However, most of them rely on a strong assumption that the entire body of the pedestrian is available, however, this is not always the case in practice. 
As shown in Fig.~\ref{fig:motivation}(a), in realistic Re-ID systems, people are always occluded by some obstacles especially in crowded places such as malls, railway stations and airports. 
Thus, it is necessary to study the occluded person Re-ID problem \cite{zhuo2018occluded}.

There are two main challenges for the occluded person Re-ID problem. Firstly, as shown in Fig.~\ref{fig:motivation}(a), occlusions always bring noise during feature extraction and feature matching. Secondly, as shown in Fig.~\ref{fig:motivation}(b), the pedestrian information in the occluded regions is always lost, making the extracted features not discriminative anymore. 
Recently, some occluded person Re-ID methods have been proposed \cite{gao2020pose,miao2019pose,he2019foreground,sun2019perceive} for the first challenge. 
They try to use key-points \cite{gao2020pose,miao2019pose} or probability maps \cite{he2019foreground,sun2019perceive} for feature alignment and increase the robustness of the representations. 
A recent work \cite{wang2020high} views key-points as nodes to construct a graph for learning the human-topology information and has proved the effectiveness of the graph for solving the occluded person Re-ID problem.
For the second challenge, some researches \cite{iodice2018partial,jin2020semantics,zhou2018aware} attempt to predict the occluded parts in the images with GANs. However, occluded regions generation is not so convincing, especially when the occlusion is serious. Therefore, the performance gains of these methods are very limited.

In this work, we propose a novel framework named Feature Recovery Transformer (FRT) to address the two challenges simultaneously. 
Specifically, the proposed framework mainly consists of three phases, semantic feature extraction, visibility graph matching, and occluded feature recovery. 
In the first phase, we employ the pose information to extract semantic features ($i.e.$, global feature, head feature, torso feature and leg feature) and calculate the visibility score for the corresponding regions. 
In the second feature matching phase, we construct a directional graph with each node containing the semantic features from the same part within a pair of images.
The weights of edges are determined by the visibility score of the starting node. 
As shown in Fig.~\ref{fig:motivation}(a), by
\IEEEpubidadjcol
promoting the flow of messages in the shared regions (which regions are visible in both images), the visibility graph pays more attention to these areas during feature matching, which is not sensitive to the occlusions. 
Based on the similarity computed by visibility graph matching, we readily obtain its $k$-nearest neighbors in the gallery for each query.
In the third occluded feature recovery phase, as shown in Fig.~\ref{fig:motivation}(d), different from other methods \cite{iodice2018partial,jin2020semantics,zhou2018aware} which use GANs to predict the occluded parts, we propose a Feature Recovery Transformer (FRT) to exploit the pedestrian information in its $k$-nearest neighbors features for occluded feature recovery. FRT considers the local information of each semantic feature in the $k$-nearest neighbors, including position, visibility score and similarity between the query. By this way, FRT is able to filter out noise in the $k$-nearest neighbors features and exploit valuable information to recover the occluded query feature. Finally, the recovered query feature is used for person Re-ID. Extensive experiments validate the consistent superiority of our FRT over prior state-of-the-art methods. Specifically, FRT outperforms state-of-the-art results by at least 6.2\% Rank-1 accuracy and 7.2\% mAP scores on the challenging Occluded-Duke dataset \cite{miao2019pose}.


The main contributions of this paper are summarized as follows: 
\begin{itemize}
\item We propose a Feature Recovery Transformer (FRT) to exploit the pedestrian information in the features of $k$-nearest neighbors for occluded feature recovery. Compared to other methods \cite{iodice2018partial,jin2020semantics,zhou2018aware} which use GANs to predict the occluded parts, our approach is more convincing and could bring much more improvements to the occluded person Re-ID performance.
\item We propose a novel visibility graph to learn the human-topology information among body parts, which is able to promote the information in the shared regions and suppress the noisy message in the occluded parts.
\item Extensive experiments on occluded, partial and holistic Re-ID datasets validate the effectiveness of our method for solving occluded person Re-ID problem.
\end{itemize}

\section{Related Work}
\textbf{Occluded and Partial Person Re-identification.} Occluded \cite{miao2019pose} and partial person re-identification \cite{Zheng_2015_ICCV} aim to find the same person, who is occluded or partially detected in the query image, from dis-joint cameras. These two problems are usually studied as the same issue in research. There are two main challenges for the occluded person Re-ID problem. Firstly, occlusions always bring noise during feature extraction and feature matching. Secondly, the pedestrian information in the occluded regions is always lost, making the extracted features not discriminative anymore. Recently, some methods have been proposed for the first challenge \cite{miao2019pose,he2018deep,he2019foreground,sun2019perceive,wang2020high,zheng2021pose,OAMN,yang}. Miao et al. \cite{miao2019pose} propose a feature alignment method based on the semantic key-points. In addition, they design a matching strategy to calculate the distance of representations in an unoccluded region. He et al. \cite{he2018deep} propose a reconstruction method for soft feature alignment and further introduce foreground-background mask to avoid the influence of backgrounds in \cite{he2019foreground}. Sun et al. \cite{sun2019perceive} propose a Visibility-aware Part Model (VPM) to learn to perceive the visibility of regions through self-supervision. Wang et al. \cite{wang2020high} utilize GCN, which considers different key-points as nodes, to embed the high-order information between various body parts. Zheng et al. \cite{zheng2021pose} propose a Guided Feature Learning with Knowledge Distillation (PGFL-KD) network to learn aligned representations of different body parts. Benefiting from the knowledge distillation and interaction-based learning, the pose estimator could be discarded in testing. Chen et al. \cite{OAMN} propose an Occlusion Aware Mask Network (OAMN), which incorporates an attention-guided mask module to extract features of body parts precisely regardless of the occlusion. Yang et al. \cite{yang} propose to discretize pose information to the visibility label of body parts for reducing the interference of noisy pose information in the occluded Re-ID problem. Zhang et al. \cite{zhang2019densely} and Jia et al. \cite{jia2022learning} attempt to  extract semantically aligned features and eliminate occlusion noises for solving the occluded Re-ID problem. Tan et al. \cite{tan2022mhsa} propose a Multi-Head Self-Attention Network (MHSA-Net) to prune noise and capture key local information from images for occluded Re-ID.
Despite the promising results achieved in occlude Re-ID, all these methods ignore the lost pedestrian information in the occluded regions. To solve this problem, \cite{iodice2018partial,jin2020semantics,zhou2018aware} attempt to predict the occluded parts in the images with GANs for the occlusion recovery. However, occluded regions generation is not so convincing, especially when the occlusion is serious. Therefore, the performance gains of these methods \cite{iodice2018partial,jin2020semantics,zhou2018aware} are very limited. Although Hou et al. \cite{hou2019} propose a Spatio-Temporal Completion network (STCnet) for recovering the appearance of the occluded parts with spatial and temporal information for video person reid, temporal information is unavailable in image-based occluded Re-ID. Different from previous methods, our method attempts to filter out noise in the $k$-nearest neighbors and fuse the query feature with valuable information in the $k$-nearest neighbors for occluded feature recovery.

\textbf{Transformer.} Vaswani et al. \cite{vaswani2017attention}
propose the Transformer to dispense the recurrence and convolutions involved in the encoding step entirely. Transformer only relies on attention mechanisms to capture the global relations between input and output for transduction problems such as machine translation and language modeling \cite{bert,bert2,bert3}. Some methods have tried to exploit Transformer in computer vision tasks such as image processing \cite{tans_img}, object detection \cite{trans_object} , semantic segmentation \cite{trans_seg}, feature matching \cite{sarlin2020superglue}, etc. For example, Chen et al. \cite{tans_img} propose Image Processing Transformer (IPT) for utilizing large-scale pre-training and achieves the state-of-the-art performance on several image processing tasks like denoising, de-raining and super-resolution. Sarlin et al. \cite{sarlin2020superglue} incorporate Transformer to establish pointwise correspondences between a pair of images for feature matching. They utilize self- (intra-image) and cross- (inter-image) attention to simulate the procedure that humans look back-and-forth at two images when matching them. Recently, Li et al. \cite{PAT} try to solve the occluded Re-ID problem by the transformer encoder-decoder architecture and propose a Part-Aware Transformer (PAT). PAT works on precisely extracting features of visible body parts by a pixel context based transformer encoder and a part prototype based transformer decoder. Different from \cite{PAT}, we utilize Transformer to exploit the pedestrian information in the $k$-nearest neighbors for recovering the occluded query features.

\textbf{Graph Convolutional Network.} Graph Convolutional Networks (GCN) is firstly proposed in \cite{scarselli2008graph} to build the relationship between graph nodes, and has been proved to be effective in many computer vision tasks \cite{gao2019graph,wang2019zero,yun2019graph}. Recently, Re-ID methods combined with GCN have also been explored \cite{cheng2018deep,wang2020high,yan2019learning}. Wang et al. \cite{wang2020high} construct the graph based on the visibility of key-points intra image, and take advantage of the affinities between various key-points. Cheng et al. \cite{cheng2018deep} formulate the structured distance into the graph Laplacian form to consider the relationships among training samples. Yan et al. \cite{yan2019learning} attempt to solve the person search by considering the context information with GCN.
\begin{figure*}[t]
\centering
\begin{overpic}[scale=0.47]{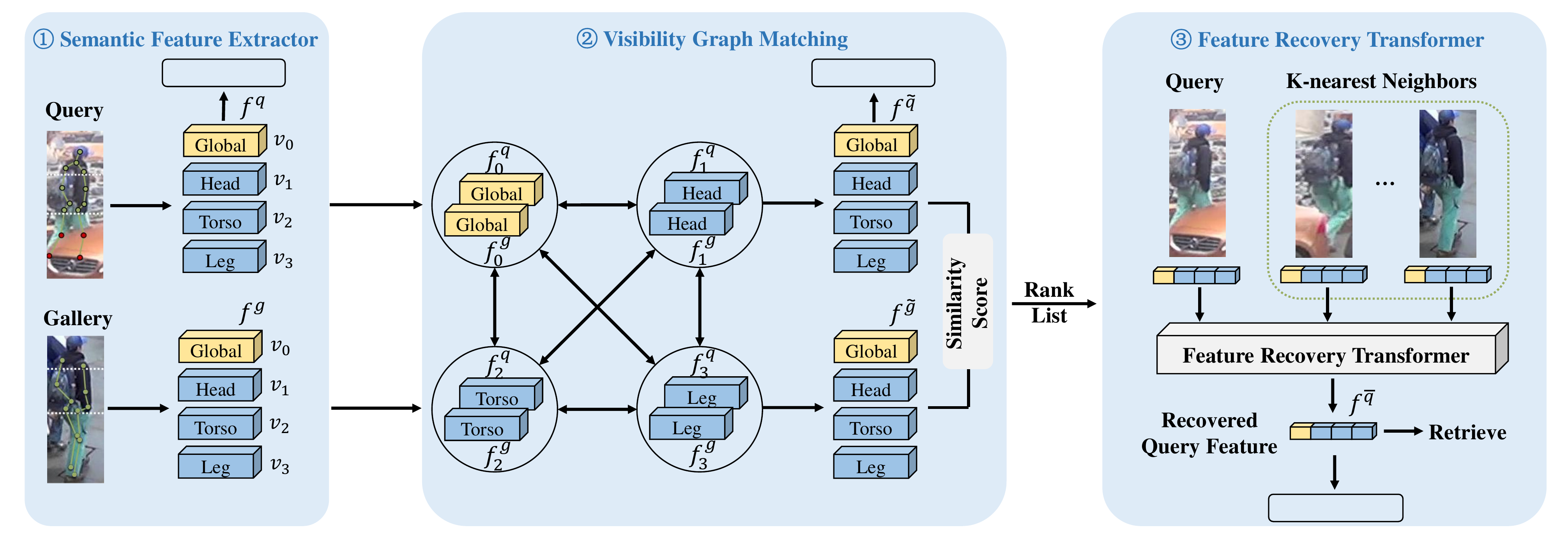}
	\put(10.8,29.4){\scriptsize{$L_\mathcal{E}$ in Eq.(\ref{loss1})}}
	\put(51.8,29.4){\scriptsize{$L_\mathcal{G}$ in Eq.(\ref{loss2})}}
	\put(81.3,1.6){\scriptsize{$L_\mathcal{T}$ in Eq.(\ref{loss:3})}}
\end{overpic}
\caption{Overview of the proposed framework. It consists of semantic feature extractor $\mathcal{E}$, visibility graph matching $\mathcal{G}$ and feature recovery transformer $\mathcal{T}$. In $\mathcal{E}$ we utilize key-points to extract semantic features and calculate the visibility score $\{v_i\}_{i=0}^3$ for each part. In $\mathcal{G}$, we input the features extracted by $\mathcal{E}$ and consider the same semantic features within a pair of images as nodes of a graph to calculate the similarity scores. According to the feature matching by $\mathcal{G}$, a rank list is produced for each query. In $\mathcal{T}$, we input the query feature and features of its $k$-nearest neighbors for recovering the occluded query feature. The recovered query feature is then utilized for retrieving.}
\label{fig:model}
\end{figure*}

\section{Our Approach}
This section introduces our proposed framework, including 1) Semantic feature extractor ($\mathcal{E}$) to extract the semantic features with pose assistance and calculate visibility scores for them; 2) Visibility graph matching ($\mathcal{G}$) to promote the information in the shared regions and learn the similarity; 3) Feature recovery transformer ($\mathcal{T}$) to recover the occluded query features with pedestrian information in the features of $k$-nearest neighbors. An overview of the proposed method is shown in Fig.~\ref{fig:model}.
\subsection{Semantic Feature Extractor}
The semantic feature extractor ($\mathcal{E}$) is demonstrated in Fig.~\ref{fig:model}. The module $\mathcal{E}$ is inspired by two cues. Firstly, part-based models have been proved to be effective for person re-identification task as they can employ both global and fine-grained local features \cite{sun2018beyond}. Secondly, 
occlusions always cause spatial misalignment during feature matching. Therefore, 
accurate feature alignment is necessary for occluded Re-ID \cite{he2018deep,he2019foreground,sun2019perceive}. Following the ideas above, we use HR-Net \cite{sun2019deep} pre-trained on the COCO dataset \cite{lin2014microsoft} for pose estimation. The model predicts 12 key-points, including shoulders, elbows, wrists, hips, knees and ankles. We exploit the pose information to divide the person image into three parts: head part, torso part and leg part. Then, the local features of these three parts together with the global feature are extracted for alignment. Additionally, we calculate the visibility score for each part as follows: 
\begin{equation}
v_{i}=\frac{\sum_{s_n\in \mathbb{R}_i}s_{n}}{\lvert \mathbb{R}_i \rvert},i=0,1,2,3,
\label{visibility}
\end{equation}
where $v_0,v_1,v_2,v_3$ are the visibility scores for the global, head, torso and leg regions respectively. $\{\mathbb{R}_i\}_{i=0}^3$ is the set of key-points in the region $i$ and $\lvert \cdot \rvert$ denotes the number of elements in the set. $s_n$ is the confidence score calculated by the pose estimator for the $n$-th key-point. The visibility score is calculated by the average confidence scores for all the key-points in the corresponding region. For example, we firstly use pose estimator to detect 6 key-points in the torso region and calculate their confidence scores by the pose estimator. Then the visibility score of the torso region is calculated by the average confidence scores for these 6 key-points. Furthermore, we set a threshold $\delta$ to determine whether the region $\mathbb{R}_i$ is completely occluded. If the visibility score $v_i$ is smaller than the $\delta$, we would regard the region $\mathbb{R}_i$ as fully occluded and set the feature of region $\mathbb{R}_i$ to zero.

\textbf{Training Loss.} To train the module $\mathcal{E}$, we use cross-entropy loss $\mathcal{L}^{\mathcal{E}}_{cross}$ and triplet $\mathcal{L}^{\mathcal{E}}_{tri}$ for all the semantic features (i.e. global, head, torso and leg feature) as follows:

\begin{equation}
\mathcal{L}^{\mathcal{E}}_{cross}=-\sum_{j=1}^{N}\log\frac{\exp(W_{y_j}^{\mathcal{E}}f_j+b^{\mathcal{E}}_{y_j})}{\sum_{k=1}^C\exp(W^{\mathcal{E}}_kf_j+b^{\mathcal{E}}_k)}, 
\label{corssloss}
\end{equation}

\begin{equation}
\mathcal{L}^{\mathcal{E}}_{tri} = \lvert \theta_{\mathcal{E}} + d_{f_j^q,f_j^P}-d_{f_j^q,f_j^N} \rvert _+,
\label{triloss}
\end{equation}

\begin{equation}
\mathcal{L_E} = \mathcal{L}^{\mathcal{E}}_{cross}+ \mathcal{L}^{\mathcal{E}}_{tri},
\label{loss1}
\end{equation}
where $N$ is the number of images in a mini-batch, $C$ is the number of classes and $y_j$ is the label for the feature $f_j$. $W_k$ and $b_k$ are the weights and bias of classifier for the $k$-th class, respectively.
$d_{f_j^q,f_j^P}$ and $d_{f_j^q,f_j^N}$ are the distance between a positive pair $(f_j^q,f_j^P)$ from the same identity and a negative pair $(f_j^q,f_j^N)$ from different identities, respectively. $\theta_{\mathcal{E}}$ is a hyper-parameter to control the margin between the negative and positive pairs in feature space. Especially, we only utilize the $\mathcal{L_E}$ to monitor the feature of the part that is not completely occluded. The reason for this is that monitoring the feature of the completely occluded regions, which is manually set to zero, would confuse the classifiers. 

\subsection{Visibility Graph Matching}
Although we have obtained the aligned pedestrian representations, occluded Re-ID is still challenging due to the interference of the occlusions during feature matching. Thus, it is necessary to suppress the meaningless message of occluded parts and enhance the meaningful features of shared regions. We resort to the graph convolutional network (GCN) \cite{wu2019simplifying} which is effective in message propagation and aggregation. As shown in Fig.~\ref{fig:model}, given two images $q$ and $g$, their feature maps 
$\{f_i^q\}_{i=0}^3$and $\{f_i^g\}_{i=0}^3$ along with their corresponding visibility scores $\{v_i^q\}_{i=0}^3$ and $\{v_i^g\}_{i=0}^3$ could be extracted and calculated by the module $\mathcal{E}$ above. We aim to construct a graph which could focus on the shared regions when matching features.

\textbf{Visibility Graph Building.}
In particular, considering a graph $\mathcal{G}=\{\mathcal{V},\mathcal{E}\}$, which consists of 4 vertices $\mathcal{V}$ and a set of edges $\mathcal{E}$ as shown in Fig.~\ref{fig:model}, we assign corresponding features 
$\{f_i^q,f_i^g\}_{i=0}^3$ to each node. We use \textbf{A} $\in \mathbb{R}^{4\times4}$ to denote the adjacent matrix. The adjacent matrix \textbf{A} is set as follows:

\begin{equation}
\begin{small}
A_{i,j}=
\begin{cases}
1,& i=j\\
\varphi_i+\mathbbm{1}(\varphi_i-\Gamma)[1-cosine(f_i^q,f_i^g)],& otherwise
\end{cases}
\end{small}
\label{A}
\end{equation}

\begin{equation}
\varphi_i = \min \{v_i^q,v_i^g\},
\label{com_visibility}
\end{equation}
where $A_{i,j}$ indicates the information propagated from node $i$ to node $j$, $\varphi_i$ denotes the shared visibility of $i$ \emph{th} part, $\Gamma$ is a margin and $\mathbbm{1}(\cdot)$ is the indicator function. A high valued $\varphi_i$ denotes that part $i$ is a shared region while a small valued $\varphi_i$ means that at least one of $i$ \emph{th} parts of the image $q$ and $g$ is occluded. The first term $\varphi_i$ in Eq.~\ref{A} indicates that the lower the shared visibility $\varphi_i$ is, the less information of node $i$ is spread. The second term $\mathbbm{1}(\varphi_i-\Gamma)[1-cosine(f_i^q,f_i^g)]$ indicates that when the shared visibility $\varphi_i$ is larger than $\Gamma$, the greater the difference between $f_i^q$ and $f_i^g$, the more their message will be propagated. This is designed to enhance the comparisons of the shared regions.

\textbf{Message Propagation.} Following GCN \cite{sarlin2020superglue}, we denote $m_i^{(l)}$ the message aggregated from all nodes to the $i$ \emph{th} node at layer $l$, which can be defined as:
\begin{equation}
m_i^{(l)}=\sigma(\widehat{A}f_i^{(l)}W_m^{(l)}),
\end{equation}
where $\widehat{A}$ is the normalized adjacency matrix, $W_m^{(l)}$ is a learnable parameter matrix, $f_i^{(l)}$ represents an element from \{$f_i^{q(l)}, f_i^{g(l)}$\} and $\sigma$ is the ReLU activation function. Furthermore, we then use the residual message passing to update all the nodes by:
\begin{equation}
f_i^{(l+1)}=f_i^{(l)}+W_r^{(l)}[f_i^{(l)},m_i^{(l)}],
\end{equation}
where $[\cdot,\cdot]$ denote concatenation and $W_r^{(l)}$ is a learnable parameter matrix.

\textbf{Feature Matching.}
After message propagation, we obtain the updated features $f^{\tilde{q}}$ and $f^{\tilde{g}}$. Then, the cosine distance is used to calculate the similarity score as follows:
\begin{equation}
s_{q,g}=cosine(f^{\tilde{q}}, f^{\tilde{g}}).
\end{equation}
According to the feature matching by $\mathcal{G}$, a rank list is produced for each query.

\textbf{Training Loss.} We use triplet and classification losses to monitor module $\mathcal{G}$ as in Eq.~\ref{loss2}. The definition of $\mathcal{L}^{\mathcal{G}}_{cross}$ and $\mathcal{L}^{\mathcal{G}}_{tri}$ is the same as Eq.~\ref{corssloss} and Eq.~\ref{triloss} respectively.
\begin{equation}
\mathcal{L_G} =  \mathcal{L}^{\mathcal{G}}_{cross}+ \mathcal{L}^{\mathcal{G}}_{tri}.
\label{loss2}
\end{equation}

\begin{figure}[t]
 \centering
 \includegraphics[width=\linewidth]{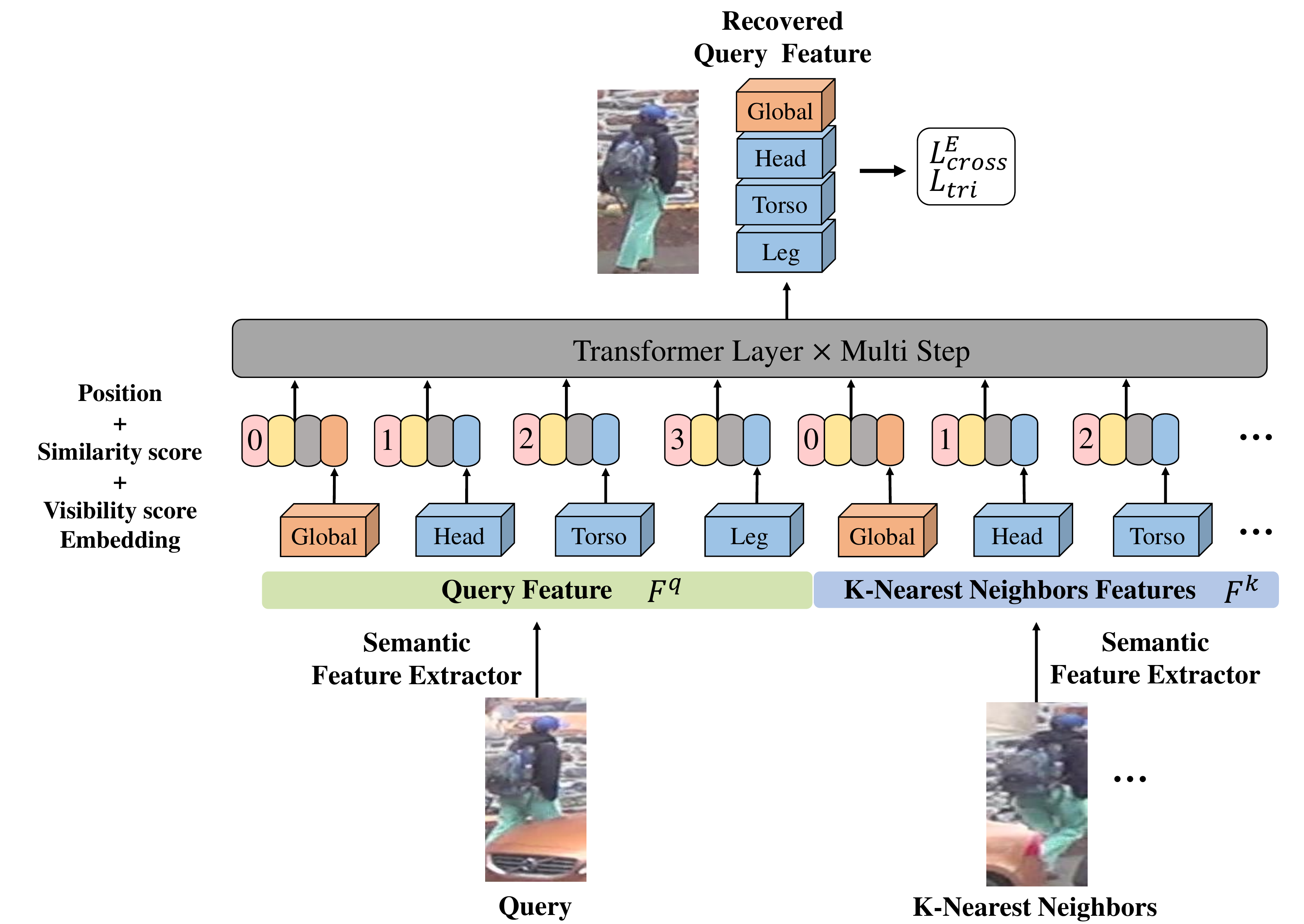}
 \caption{Illustration of the proposed feature recovery transformer ($\mathcal{T}$). $F^q$, $F^k$ are query feature and $k$-nearest neighbors features respectively.}
 \label{transformer}
\end{figure}
\subsection{Feature Recovery Transformer} 
Even so, learned features still suffer from loss of pedestrian information caused by the occlusions. To solve this problem, we propose feature recovery transformer ($\mathcal{T}$) for recovering the occluded features.
In the gallery, there possesses lots of pedestrian information, which hides the cues about complete features recovery. Inspired by the success of Transformer \cite{vaswani2017attention} and local feature matching \cite{sarlin2020superglue}, we want to employ the attention mechanism of Transformer to aggregate the pedestrian information in the $k$-nearest neighbors features for recovering the occluded query feature. 

Although both the Feature Recovery Transformer and re-ranking strategy \cite{zhong2017re} utilize the $k$-nearest neighbors information, we emphasize that the motivation and implementation are different. Re-ranking \cite{zhong2017re} works on re-calculating the distance in the k-nearest neighbors to re-rank the retrieval results. Feature Recovery Transformer focuses on recovering the occluded query feature by filtering out noise in the $k$-nearest neighbors and fusing the query feature with valuable information in the $k$-nearest neighbors features.

\textbf{Local Information Embedding.} The feature recovery transformer is illustrated in Fig.~\ref{transformer}. We input the concatenation of query feature and its $k$-nearest neighbors features to the $\mathcal{T}$. For each part feature $f$, we embed its position, similarity score between the query and visibility score into a high-dimensional vector with a Multilayer Perceptron (MLP) as:
\begin{equation}
f=f+MLP(p,cos,v),
\end{equation}
where $p$ is the position for $f$, $p=0,1,2,3$ stand for the global, head, torso and leg part respectively, $cos$ is the cosine distance between the pedestrian feature and query feature and $v$ is the visibility score for $f$. This embedding enables $\mathcal{T}$ to consider the local information of each part feature during the query feature recovery. 

\textbf{Transformer Layer.} The transformer layer works on aggregating the information from $k$-nearest neighbors features to recover query feature. Inspired by Transformer \cite{vaswani2017attention}, the key, query and value can be calculated by:
\begin{equation}
\begin{aligned}
q^{(l)}&=W_1^{(l)}f^{q(l)}+b_1^{(l)},\\
\begin{bmatrix}k\\s\end{bmatrix}^{(l)}&=\begin{bmatrix}W_{2}\\W_{3}\end{bmatrix}^{(l)}f^{k}+\begin{bmatrix}b_2\\b_3\end{bmatrix}^{(l)},
\end{aligned}
\end{equation}
where, $q, k, s, f^q, f^k$ indicate query, key, value, query feature and $k$-nearest neighbors features respectively and $l$ is the layer number. Each layer $l$ has its own learnable projection matrix, shared for all part features $f^q$ and $f^k$. Then, the message propagated to $f^q$ is computed as:
\begin{equation}
m_{f^{t(l)} \to f^{q(l)}} = \sum_{f^k\in F^k} Softmax(q^{{(l)}\top}k^{(l)})s^{(l)}.
\end{equation}
The final projection is a linear projection:
\begin{equation}
f^{\overline{q}}=W_{final}f^{q(L)}+b,
\end{equation}
where $f^{\overline{q}}$ is the recovered query representation for better retrieval. It is worth noting that, the $k$-nearest neighbors features are not updated during the whole process. Restricted by pages, we do not show more details of Transformer, please refer to the paper \cite{sarlin2020superglue,vaswani2017attention}.
In addition, we employ multi-step mechanism as follows:
\begin{equation}
f^{\overline{q}(s)}=\mathcal{T}_s\mathcal{T}_{s-1}...\mathcal{T}_0(f^{\overline{q}(0)},f^{\overline{q}(0)}),
\label{multi}
\end{equation}
where $s$ denotes conduct $\mathcal{T}$ for $s$ times. The recovered query feature $f^{\overline{q}(s)}$ is then utilized for retrieval.

\textbf{Training Loss.} The $\mathcal{T}$ only recover the query features and would leave the $k$-nearest neighbors features unchanged. Therefore, we use the classifier $\mathcal{L
}^\mathcal{E}_{cross}$ designed for $\mathcal{E}$ to train the $\mathcal{T}$, and freeze $\mathcal{L
}^\mathcal{E}_{cross}$ during the whole training process of $\mathcal{T}$ to ensure that the gallery features and recovered query features are in the same feature space. The training loss of the module $\mathcal{T}$ can be defined as:
\begin{equation}
\mathcal{L_T} =  \mathcal{L}^{\mathcal{E}}_{cross}+ \mathcal{L}^{\mathcal{T}}_{tri},
\label{loss:3}
\end{equation}
where $\mathcal{L}^{\mathcal{E}}_{cross}$ is the classifier designed for the module $\mathcal{E}$ and the definition of $\mathcal{L}^{\mathcal{T}}_{tri}$ is the same as Eq.~\ref{triloss}.

\section{Experiments}
\subsection{Setup}
\textbf{Datasets.} We evaluate our method on the following six datasets and compare it with the state-of-the-art methods, the details of the datasets are illustrated in Table~\ref{table:data}. 
1) The Occluded-Duke dataset \cite{miao2019pose} is derived from DukeMTMC-reID \cite{ristani2016performance} by filtering out some overlap pictures and leaving occluded images. It consists of $15,618$ training images, $2,210$ occluded query images and $17,661$ images in gallery. 2) The Occluded-ReID dataset \cite{zhuo2018occluded} contains $1000$ occluded query images and $1000$ full-body gallery pictures. 3) The Partial-REID dataset \cite{zheng2015partial} includes $600$ images from 60 people in test set, with five full-body images and five partial images for each person. 4) Partial-iLIDS dataset \cite{he2018deep} is selected from iLIDS \cite{zheng2011person}. It contains 238 occluded images from 119 people captured in the airport. Specifically, the Occluded-ReID \cite{zhuo2018occluded}, Partial-REID \cite{zheng2015partial} and Partial-iLIDS dataset \cite{he2018deep} only contain test sets, following \cite{he2019foreground,wang2020high}, the Market-1501 \cite{zheng2015scalable} is used for training. 5) The Market-1501 \cite{zheng2015scalable} dataset consists of 32,688 images of 1,501 subjects captured by six cameras, and only few of occluded or partial person images are included. 6) DukeMTMC-reID \cite{ristani2016performance} is a holistic dataset which contains 1,404 identities, 16,522 training images, 2,228 queries, and 17,661 gallery images.

\begin{table}[h]
\center
\setlength{\tabcolsep}{0.3mm}
\renewcommand\arraystretch{1.2}
\caption{Datasets details. We evaluate our method on 6 public datasets, including 2 occluded, 2 partial and 2 holistic ones. Occluded-ReID, Partial-ReID and Partial-iLIDS datasets adopt cross-domain setting.}
\begin{tabular}{|c|c|c|c|}
\hline
\multirow{2}{*}{Dataset} & \multicolumn{3}{c|}{Nums (ID/Image)} \\ \cline{2-4} 
& Training & Query & Gallery \\ \hline
Occluded-Duke \cite{miao2019pose} & 702/15,618 & 519/2210 & 1,110/17,661 \\ \hline
Occluded-ReID \cite{zhuo2018occluded} & - & 200/1,000 & 200/1,000 \\ \hline
Partial-REID \cite{zheng2015partial} & - & 60/300 & 60/300 \\ \hline
Partial-iLIDS \cite{he2018deep} & - & 119/119 & 119/119 \\ \hline
Market-1501 \cite{zheng2015scalable} & 751/12,936 & 750/3,368 & 750/19,732 \\ \hline
DukeMTMC-reID \cite{ristani2016performance} & 702/16,522 & 702/2,228 & 1,110/17,661 \\ \hline
\end{tabular}
\label{table:data}
\end{table}

\begin{table*}[]
\center
\caption{Performance (\%) comparisons with the state-of-the-art methods on the two occluded datasets, $i.e.,$ Occluded-Duke  \cite{miao2019pose} and Occluded-ReID \cite{zhuo2018occluded}. Our method achieves the best performance on the occluded-Duke dataset. The best performance is highlighted in bold.}
\setlength{\tabcolsep}{2.05mm}
\renewcommand\arraystretch{1.3} 
\begin{tabular}{|c|c|cc|cc|cc|}
\hline
 \multirow{2}{*}{Methods} & \multirow{2}{*}{Reference} & \multicolumn{2}{c|}{Occluded-Duke} &
\multicolumn{2}{c|}{Occluded-REID} &
\multicolumn{2}{c|}{Average}\\ \cline{3-8} 
                                                                   &                            & \multicolumn{1}{c|}{Rank-1} & mAP  &
                                          \multicolumn{1}{c|}{Rank-1} & mAP  &
                                          \multicolumn{1}{c|}{Rank-1} & mAP\\ \hline
 Part-Aligned \cite{zhao2017deeply}            & ICCV 2017                  & \multicolumn{1}{c|}{28.8}   & 20.2 &
\multicolumn{1}{c|}{-}   & - &
\multicolumn{1}{c|}{-}   & -\\
                                           PCB \cite{sun2018beyond}                     & ECCV 2018                  & \multicolumn{1}{c|}{42.6}   & 33.7 &
\multicolumn{1}{c|}{41.3}   & 38.9 &
\multicolumn{1}{c|}{42.0}   & 36.3\\ \hline
 Part Bilinear \cite{suh2018part}           & ECCV 2018                  & \multicolumn{1}{c|}{36.9}   & -    &
\multicolumn{1}{c|}{-}   & - &
\multicolumn{1}{c|}{-}   & -\\
                                           FD-GAN \cite{ge2018fd}                  & NIPS 2018                  & \multicolumn{1}{c|}{40.8}   & -    &
\multicolumn{1}{c|}{-}   & - &
\multicolumn{1}{c|}{-}   & -\\ \hline
 DSR \cite{he2018deep}                     & CVPR 2018                  & \multicolumn{1}{c|}{40.8}   & 30.4 &
\multicolumn{1}{c|}{72.8}   & 62.8 &
\multicolumn{1}{c|}{56.8}   & 46.6\\ \hline
Ad-Occluded \cite{huang2018adversarially}             & CVPR 2018                  & \multicolumn{1}{c|}{44.5}   & 32.2 &
\multicolumn{1}{c|}{-}   & - &
\multicolumn{1}{c|}{-}   & -\\
                                           PGFA  \cite{miao2019pose}                   & ICCV 2019                  & \multicolumn{1}{c|}{51.4}   & 37.3 &
\multicolumn{1}{c|}{-}   & - &
\multicolumn{1}{c|}{-}   & -\\
                                           HOReID \cite{wang2020high}                  & CVPR 2020                  & \multicolumn{1}{c|}{55.1}   & 43.8 &
\multicolumn{1}{c|}{80.3}   & 70.2 &
\multicolumn{1}{c|}{67.7}   & 57.0\\
                                          PVPM \cite{PVPM}      & CVPR 2020                  & \multicolumn{1}{c|}{47.0}   & 37.7 &
\multicolumn{1}{c|}{70.4}   & 61.2 &
\multicolumn{1}{c|}{58.7}   & 49.5\\
                                           Pirt \cite{PIRT}                    & ACM MM 2021                & \multicolumn{1}{c|}{60.0}   & 50.9 &
\multicolumn{1}{c|}{-}   & - &
\multicolumn{1}{c|}{-}   & -\\
                                           PGFA-KD \cite{PGFA-KD}                 & ACM MM 2021                & \multicolumn{1}{c|}{63.0}   & 54.1 &
\multicolumn{1}{c|}{80.7}   & 70.3 &
\multicolumn{1}{c|}{71.9}   & 62.2\\
                                          Yang et al. \cite{yang}             & ICCV 2021                  & \multicolumn{1}{c|}{62.2}   & 46.3 &
\multicolumn{1}{c|}{81.0}   & 71.0 &
\multicolumn{1}{c|}{71.6}   & 58.7\\
                                           OAMN \cite{OAMN}                     & ICCV 2021                  & \multicolumn{1}{c|}{62.6}   & 46.1 &
\multicolumn{1}{c|}{-}   & - &
\multicolumn{1}{c|}{-}   & -\\
                                           PAT \cite{PAT}                     & CVPR 2021                  & \multicolumn{1}{c|}{64.5}   & 53.6 &
\multicolumn{1}{c|}{\textbf{81.6}}   & \textbf{72.1} &
\multicolumn{1}{c|}{73.1}   & 62.9\\ \cline{1-8}
                                           FRT (ours)               &                            & \multicolumn{1}{c|}{\textbf{70.7}}   & \textbf{61.3} &
\multicolumn{1}{c|}{80.4}   & 71.0 &
\multicolumn{1}{c|}{\textbf{75.6}}   & \textbf{66.2}\\\hline
\end{tabular}
\label{table:occluduke}
\end{table*}

\begin{table*}[]
\center
\caption{Performance (\%) comparisons with the state-of-the-art methods on the two partial datasets, $i.e.,$ Partial-REID \cite{zheng2015partial} and Partial-iLIDS \cite{he2018deep}. Our method achieves the best performance on the Partial-REID \cite{zheng2015partial}. The best performance is highlighted in bold.}
\setlength{\tabcolsep}{1.8mm}
\renewcommand\arraystretch{1.3} 
\begin{tabular}{|c|c|cc|cc|cc|}
\hline
\multirow{2}{*}{Methods} & \multirow{2}{*}{Reference} &  \multicolumn{2}{c|}{Partial-REID}    & \multicolumn{2}{c|}{Partial-iLIDS}    & \multicolumn{2}{c|}{Average}   \\ \cline{3-8} 
                         &                            &  \multicolumn{1}{c|}{Rank-1} & Rank-3 & \multicolumn{1}{c|}{Rank-1} & Rank-3 & \multicolumn{1}{c|}{Rank-1} & Rank-3 \\ \hline
DSR \cite{he2018deep}                     & CVPR 2018                  & \multicolumn{1}{c|}{58.8}   & 67.2   & \multicolumn{1}{c|}{50.7}   & 70.0   & \multicolumn{1}{c|}{54.8}   & 68.6      \\
AFPB \cite{zhuo2018occluded}                     & ICME 2018                  & \multicolumn{1}{c|}{78.5}   & -      & \multicolumn{1}{c|}{-}   & -& \multicolumn{1}{c|}{-}   & -            \\
FPR \cite{he2019foreground}                     & ICCV 2019                   & \multicolumn{1}{c|}{68.1}   & -      & \multicolumn{1}{c|}{81.0}   & -& \multicolumn{1}{c|}{74.6}   & -            \\
PGFA \cite{miao2019pose}                    & ICCV 2019                      & \multicolumn{1}{c|}{69.1}   & 80.9   & \multicolumn{1}{c|}{68.0}   & 80.0& \multicolumn{1}{c|}{68.6}   & 80.5         \\
VPM \cite{sun2019perceive}                     & CVPR 2019                     & \multicolumn{1}{c|}{65.5}   & 74.8   & \multicolumn{1}{c|}{67.7}   & 81.9& \multicolumn{1}{c|}{66.6}   & 78.4         \\
STNReID  \cite{stnreid}                 & TMM 2020                   & \multicolumn{1}{c|}{66.7}   & 80.3   & \multicolumn{1}{c|}{54.6}   & 71.3& \multicolumn{1}{c|}{60.7}   & 75.8         \\
PVPM  \cite{PVPM}                 & CVPR 2020                   & \multicolumn{1}{c|}{78.3}   & 87.7   & \multicolumn{1}{c|}{-}   & -& \multicolumn{1}{c|}{-}   & -         \\
HOReID  \cite{wang2020high}                 & CVPR 2020                   & \multicolumn{1}{c|}{85.3}   & 91.0   & \multicolumn{1}{c|}{72.6}   & 86.4& \multicolumn{1}{c|}{79.0}   & 88.7         \\
PGFA-KD \cite{PGFA-KD}                 & ACM MM 2021                 & \multicolumn{1}{c|}{85.1}   & 90.8   & \multicolumn{1}{c|}{74.0}   & 86.7& \multicolumn{1}{c|}{80.0}   & 88.8         \\
OAMN \cite{OAMN}                     & ICCV 2021                      & \multicolumn{1}{c|}{86.0}   & -      & \multicolumn{1}{c|}{\textbf{77.3}}   & -& \multicolumn{1}{c|}{81.7}   & -            \\
PAT \cite{PAT}                     & CVPR 2021                   & \multicolumn{1}{c|}{88.0}   & 92.3   & \multicolumn{1}{c|}{76.5}   & \textbf{88.2}& \multicolumn{1}{c|}{\textbf{82.2}}   & \textbf{90.3}         \\ \hline
FRT (ours)               &                             & \multicolumn{1}{c|}{\textbf{88.2}}   & \textbf{93.2}   & \multicolumn{1}{c|}{73.0}   & 87.0   & \multicolumn{1}{c|}{80.6}   & 90.1      \\ \hline
\end{tabular}
\label{table:occlu}
\end{table*}

\textbf{Training Details.} Our baseline is built based on the open-source project "fastreid" \cite{fastreid}. We resize all the training images into $384 \times 128$. We set the number of feature channels $c$ to $512$ and batch size $N$ to $64$. Following the work of \cite{wang2018learning}, the global average pooling (GAP) and fully connected layers are removed from the original ResNet-50 \cite{He2016DeepRL} architecture and the stride of the last convolution layer is set to 1. The parameter $s$ in Eq.~\ref{multi} equals 3 and we input the $5$-nearest neighbors features to the $\mathcal{T}$. We exploit one GCN layer in our method.

\textbf{Evaluation Metrics.} We utilize mean average precision (mAP) and Cumulative Matching Characteristic (CMC) curves to evaluate the performance of various Re-ID models. All the experiments are conducted in a single query setting.

\subsection{Comparison with State-of-the-art Methods}
\textbf{Results on the Occluded Datasets.} In Table~\ref{table:occluduke}, we compare our method with the state-of-the-art Re-ID methods on the two occluded datasets, $i.e.,$ Occluded-Duke  \cite{miao2019pose} and Occluded-ReID \cite{zhuo2018occluded}. Four kinds of methods are compared, they are holistic methods \cite{sun2018beyond,zhao2017deeply}, key-points based methods \cite{ge2018fd,suh2018part}, partial Re-ID methods \cite{he2018deep} and occluded Re-ID methods \cite{huang2018adversarially,miao2019pose,wang2020high,PIRT,PGFA-KD,yang,OAMN,PAT,PVPM}. The result shows that FRT outperforms other methods on Occluded-Duke dataset \cite{miao2019pose}, which demonstrates the effectiveness of our FRT in dealing with the occluded Re-ID problem. Specifically, on the Occluded-Duke \cite{miao2019pose} dataset, FRT achieves the best result with Rank-1 accuracy of $70.7\%$ and mAP of $61.3\%$, which is at least $6.2\%$ and $7.7\%$ higher than the corresponding metrics of other methods. On the Occluded-REID dataset \cite{zhuo2018occluded}, FRT achieves the competitive results to PAT \cite{PAT} with $80.4 \%$ Rank-1 accuracy and $71.0 \%$ mAP.

\textbf{Results on the Partial Datasets.} In Table~\ref{table:occlu}, we compare our method with the state-of-the-art Re-ID methods on the two partial datasets, $i.e.,$ Partial-REID \cite{zheng2015partial} and Partial-iLIDS \cite{he2018deep}. Accompanied by occluded images, partial ones often occur due to outliers of camera views, imperfect detection, and so on. As we can see, our method achieves the best results on the Partial-REID dataset \cite{zheng2015partial}, which outperforms other methods by at least $0.2 \%$ Rank-1 accuracy and $0.9\%$ Rank-3 accuracy. FRT also achieves competitive results on the Partial-iLIDS dataset \cite{he2018deep}. We think there are three reasons for the less performance gains on the three small-scale occluded and partial datasets, $i.e.,$  Occluded-ReID \cite{zhuo2018occluded}, Partial-REID \cite{zheng2015partial} and Partial-iLIDS \cite{he2018deep} than on the Occluded-Duke \cite{miao2019pose}: 1) The three small-scale occluded and partial datasets $i.e.$  Occluded-ReID \cite{zhuo2018occluded}, Partial-REID \cite{zheng2015partial} and Partial-iLIDS \cite{he2018deep} adopt cross-domain setting, which utilize Market-1501 as the training set and test on the other domains. Domain bias in the cross-domain evaluation would have a negative impact on the performance of our model.
2) The gallery of the Occluded-Duke \cite{miao2019pose},  Occluded-ReID \cite{zhuo2018occluded}, Partial-REID \cite{zheng2015partial} and Partial-iLIDS \cite{he2018deep} contain 16,5,5,1 images for each identity respectively. Therefore, on Occluded-Duke [10], the feature recovery transformer is able to employ more pedestrian information in the gallery, resulting in better recovered query features and bigger Re-ID performance improvements on Occluded-Duke than other three small-scale occluded and partial datasets.
3) Occluded-Duke \cite{miao2019pose} is the largest dataset for studying the occluded Re-ID problem, and the results on the Occluded-Duke are more reliable for occluded Re-ID performance evaluation.

\begin{table*}[]
\center
\caption{Performance (\%) comparisons with the state-of-the-art Re-ID methods on holistic datasets $i.e.$ Market-1501 \cite{zheng2015scalable} and DukeMTMC-reID \cite{ristani2016performance}. Our method achieves best performance on holistic Re-ID. The best performance is highlighted in bold.}
\setlength{\tabcolsep}{1.8mm}
\renewcommand\arraystretch{1.3} 
\begin{tabular}{|c|c|c|cc|cc|cc|}
\hline
\multirow{2}{*}{}                         & \multirow{2}{*}{Methods} & \multirow{2}{*}{Reference} & \multicolumn{2}{c|}{Market-1501}   & \multicolumn{2}{c|}{DukeMTMC-reID}   & \multicolumn{2}{c|}{Average} \\ \cline{4-9} 
                                          &                          &                            & \multicolumn{1}{c|}{Rank-1} & mAP  & \multicolumn{1}{c|}{Rank-1} & mAP  & \multicolumn{1}{c|}{Rank-1} & mAP  \\ \hline
\multirow{2}{*}{Holistic Methods}         & PCB \cite{sun2018beyond}                     & ECCV 2018                  & \multicolumn{1}{c|}{92.3}   & 77.4 & \multicolumn{1}{c|}{81.8}   & 66.1 & \multicolumn{1}{c|}{87.1}   & 71.8  \\
                                          & BOT \cite{luo2019bag}                     & ICCV 2019                  & \multicolumn{1}{c|}{94.1}   & 85.7 & \multicolumn{1}{c|}{86.4}   & 76.4 & \multicolumn{1}{c|}{90.3}   & 81.1  \\ \hline
\multirow{2}{*}{Key-points Based Methods} & Part Bilinear \cite{suh2018part}            & ECCV 2018                  & \multicolumn{1}{c|}{90.2}   & 76.0 & \multicolumn{1}{c|}{82.1}   & 64.2 & \multicolumn{1}{c|}{86.2}   & 70.1   \\
                                          & FD-GAN \cite{ge2018fd}                   & NIPS 2018                  & \multicolumn{1}{c|}{90.5}   & 77.7 & \multicolumn{1}{c|}{80.0}   & 64.5 & \multicolumn{1}{c|}{85.3}   & 71.1   \\ \hline
Partial Re-ID Methods                     & DSR \cite{he2018deep}                      & CVPR 2018                  & \multicolumn{1}{c|}{83.5}   & 64.2 & \multicolumn{1}{c|}{-}      & -     & \multicolumn{1}{c|}{-}   & -  \\ \hline
\multirow{8}{*}{Occluded Re-ID Methods}   & Ad-Occluded \cite{huang2018adversarially}              & CVPR 2018                  & \multicolumn{1}{c|}{84.4}   & 66.9 & \multicolumn{1}{c|}{79.1}   & 62.1 & \multicolumn{1}{c|}{81.8}   & 64.5   \\
                                          & FPR \cite{he2019foreground}                      & ICCV 2019                  & \multicolumn{1}{c|}{95.4}   & 86.5 & \multicolumn{1}{c|}{88.6}   & 76.4 & \multicolumn{1}{c|}{92.0}   & 81.5   \\
                                          & PGFA \cite{miao2019pose}                     & ICCV 2019                  & \multicolumn{1}{c|}{91.2}   & 76.8 & \multicolumn{1}{c|}{82.6}   & 65.5  & \multicolumn{1}{c|}{86.9}   & 71.2  \\
                                          & HOReID \cite{wang2020high}                   & CVPR 2020                  & \multicolumn{1}{c|}{91.0}   & 85.3 & \multicolumn{1}{c|}{86.4}   & 72.6  & \multicolumn{1}{c|}{88.7}   & 79.0  \\
                                          & Pirt \cite{PIRT}                    & ACM MM 2021                & \multicolumn{1}{c|}{94.1}   & 86.3 & \multicolumn{1}{c|}{88.9}   & 77.6 & \multicolumn{1}{c|}{91.5}   & 82.0   \\
                                          & PGFA-KD \cite{PGFA-KD}                 & ACM MM 2021                & \multicolumn{1}{c|}{95.3}   & 87.2 & \multicolumn{1}{c|}{89.6}   & 79.5 & \multicolumn{1}{c|}{92.5}   & 83.4   \\
                                          & OAMN \cite{OAMN}                     & ICCV 2021                  & \multicolumn{1}{c|}{93.2}   & 79.8 & \multicolumn{1}{c|}{86.3}   & 72.6 & \multicolumn{1}{c|}{89.8}   & 76.2   \\
                                          & PAT \cite{PAT}                     & CVPR 2021                  & \multicolumn{1}{c|}{95.4}   & 88.0 & \multicolumn{1}{c|}{88.8}   & 78.2 & \multicolumn{1}{c|}{92.1}   & 83.1   \\ \cline{2-9}
                                          & FRT (ours)               &                            & \multicolumn{1}{c|}{\textbf{95.5}}   & \textbf{88.1} & \multicolumn{1}{c|}{\textbf{90.5}}   & \textbf{81.7}  & \multicolumn{1}{c|}{\textbf{93.0}}   & \textbf{84.9}  \\ \hline
\end{tabular}
\label{table:holistic}
\end{table*}

\textbf{Results on Holistic Datasets.} Although recent occluded/partial person Re-ID methods have made progress on occluded/partial datasets, their performances are always unsatisfying on the holistic datasets. In this part, we show that our method can also achieve comparable state-of-the-art performances on the holistic datasets Market-1501 \cite{zheng2015scalable} and DukeMTMC-reID \cite{ristani2016performance}. The results are shown in Table~\ref{table:holistic}. We compare the proposed FRT with two holistic Re-ID methods \cite{luo2019bag,sun2018beyond}, two key-points based methods \cite{suh2018part,ge2018fd}, one partial Re-ID method \cite{he2018deep} and eight occluded Re-ID methods \cite{huang2018adversarially,he2019foreground,miao2019pose,wang2020high,PIRT,PGFA-KD,OAMN,PAT}. The result shows that on the Market-1501 \cite{zheng2015scalable} our proposed FRT achieves the best results with $95.5\%$ Rank-1 accuracy and $88.1\%$ mAP and on the 
DukeMTMC-reID \cite{ristani2016performance} FRT achieves the best results with $90.5\%$ Rank-1 accuracy and $81.7\%$ mAP, which outperforms other methods by at least $0.9\%$ and $2.2\%$ respectively.

\subsection{Comparison with Post-Processing Techniques} 
\begin{figure*}[t]
 \centering
 \includegraphics[width=0.95\textwidth]{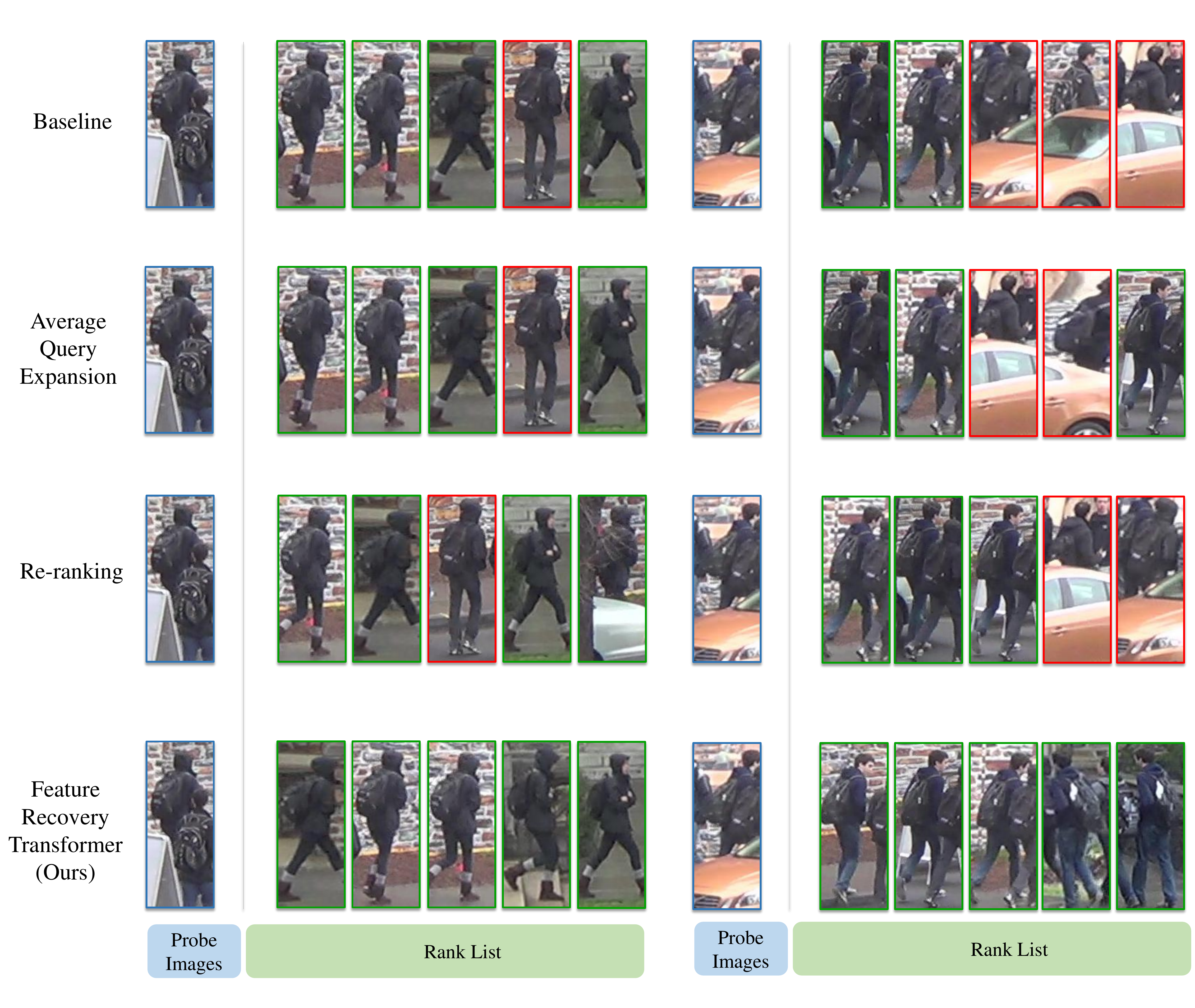}
 \caption{Visualization of the comparison of our proposed FRT and other state-of-the-art post-processing techniques, $i.e.$ average query expansion \cite{chum2007total} and re-ranking \cite{zhong2017re}. Green and red rectangles indicate correct and error retrieval results,respectively.}
 \label{fig:postvisual}
\end{figure*}

As the visibility graph and feature recovery transformer work in the feature matching stage, we additionally compare FRT with other state-of-the-art post-processing techniques, $i.e.$ re-ranking \cite{zhong2017re} and average query expansion (AQE) \cite{chum2007total} on Occluded-Duke dataset. The result is shown in Table~\ref{table:pose-process}. From the result we can see that FRT has the highest Rank-1 accuracy of $70.7\%$. We think the main reason for the higher Rank-1 accuracy of FRT is that $\mathcal{T}$ is able to filter out the noisy message in the $k$-nearest neighbors, and employ valuable information instead of weighted sum of all the features to recover the query features. However, re-ranking achieves the highest mAP of $63.7\%$, which is $2.4\%$ higher than the FRT. We think the reason for this is that re-ranking attempts to calculate the $k$-nearest neighbors for all the candidates in the rank list and recalculate the Jaccard distance for re-ranking, which is better for the mean accuracy. In addition, the result in the last row indicates that FRT and re-ranking are not conflicting, they can be integrated for better performance. We visualize the comparison of FRT, average query expansion (AQE) \cite{chum2007total} and re-ranking \cite{zhong2017re} in Fig.~\ref{fig:postvisual}.

\subsection{Further Analysis}
\textbf{Ablation Study of Proposed Modules.}
In this part, we analyze our proposed semantic feature extractor ($\mathcal{E}$), visibility graph matching ($\mathcal{G}$) and feature recovery transformer ($\mathcal{T}$) on Occluded-Duke dataset. The results are shown in Table~\ref{table:proposed modules}. Firstly, in index-1, we can see that thanks to the feature alignment by extracting semantic features with key-points, $\mathcal{E}$ is able to achieve $56.5\%$ Rank-1 accuracy and $48.6\%$ mAP on the Occluded-Duke dataset. Secondly, In index 2, affinities among different body parts are considered and the information in the shared regions is promoted. This gives $3.4\%$ and $3.2\%$ improvement to the Rank-1 accuracy and mAP respectively and demonstrates the effectiveness of $\mathcal{G}$. Thirdly, in index 1 and 3, we can see that $\mathcal{T}$ is able to give Rank-1 accuracy of $13\%$ and mAP of $11.2\%$ improvement to the $\mathcal{E}$. Finally, in index 3 and 4, $\mathcal{G}$ gives another $1.2\%$ and $1.5\%$ higher points to Rank-1 accuracy and mAP respectively to the $\mathcal{T}$.

\begin{table}[t]
\center
\caption{Performance (\%) comparisons with the state-of-the-art post-processing techniques on Occuluded-Duke dataset. $\mathcal{E}$ indicates semantic feature extractor. Our model achieves the best results and could be integrated with other post-processing techniques for better results.}
\setlength{\tabcolsep}{2.9mm}
\renewcommand\arraystretch{1.1}
\begin{tabular}{@{}|c|c|c|c|@{}}
\hline
Methods & Rank-1 & mAP \\ 
\hline
PGFA \cite{miao2019pose} & 51.4 & 37.3 \\
PGFA \cite{miao2019pose} + re-ranking & 52.4 & 46.8 \\
HOReID \cite{wang2020high} & 55.1 & 43.8 \\
HOReID \cite{wang2020high} + re-ranking & 58.3 & 49.2 \\
Pirt \cite{PIRT} & 60.0 & 50.9 \\
Pirt \cite{PIRT} + re-ranking  & 62.1 & 59.3 \\
$\mathcal{E}$ & 56.5 & 48.6 \\
$\mathcal{E}$ + AQE \cite{chum2007total}& 62.8 & 60.2 \\
$\mathcal{E}$ + re-ranking \cite{zhong2017re} & 64.6 & 63.7 \\
\hline
FRT ($ours$) & 70.7 & 61.3\\
FRT ($ours$) + re-ranking \cite{zhong2017re} & \textbf{70.8} & \textbf{65.0}\\
\hline
\end{tabular}
\label{table:pose-process}
\end{table}

\begin{table}[t]
\center
\caption{Ablation study of the proposed modules on Occuluded-Duke dataset. $\mathcal{E}$ is the semantic feature extractor, $\mathcal{G}$ is the visibility graph matching and $\mathcal{T}$ is the feature recovery transformer. The results validate the effectiveness of the three proposed modules.}
\setlength{\tabcolsep}{3.6mm}
\renewcommand\arraystretch{1.1}
\begin{tabular}{@{}|cccccc|@{}}
\hline
Index & $\mathcal{E}$ & $\mathcal{G}$ & $\mathcal{T}$ & Rank-1 & mAP \\
\hline
1 & $\surd$ & $\times$ & $\times$ & 56.5 & 48.6 \\
2 & $\surd$ & $\surd$ & $\times$ & 59.9 & 51.4 \\
3 & $\surd$ & $\times$ & $\surd$ & 69.5 & 59.8 \\
4 & $\surd$ & $\surd$ & $\surd$ & \textbf{70.7} & \textbf{61.3} \\ 
\hline
\end{tabular}
\label{table:proposed modules}
\end{table}

\begin{figure}[t]
 \centering
 \includegraphics[width=\linewidth]{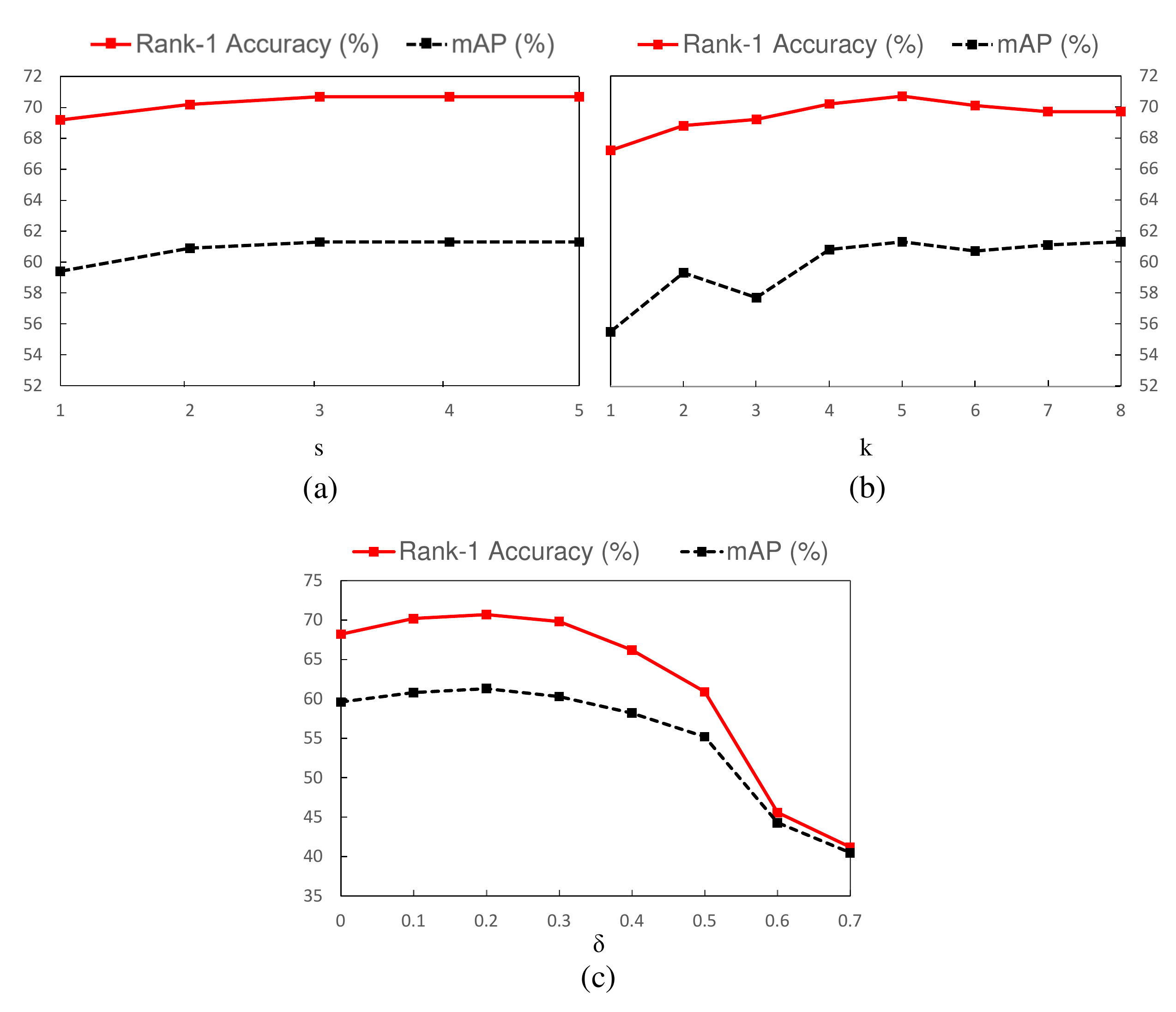}
 \caption{Analysis of parameters $s$, $k$ and $\delta$ on the Occluded-Duke dataset. $s$ indicates conduct $\mathcal{T}$ for $s$ times, $k$ indicates input the $k$-nearest neighbors features to the $\mathcal{T}$ and $\delta$ is the threshold for the semantic feature extractor.}
 \label{fig:s,k}
\end{figure}
\begin{figure}[t]
 \centering
 \includegraphics[width=\linewidth]{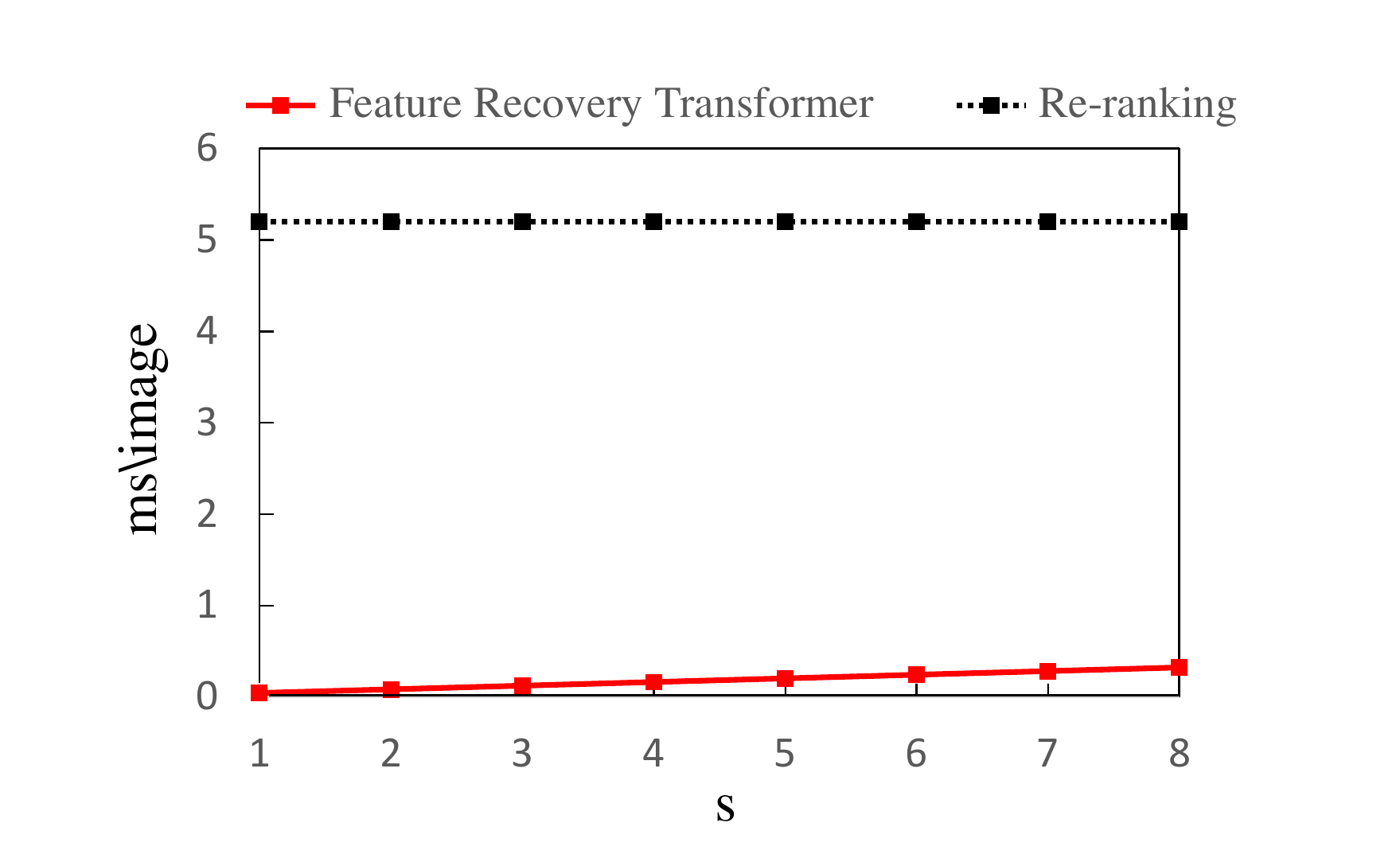}
 \caption{Analysis of
 the time consumption of the feature recovery transformer. $s$ indicates conduct $\mathcal{T}$ for $s$ times. The results show that our feature recovery transformer has less time consumption.}
 \label{fig:time}
\end{figure}

\textbf{Analysis of Parameters.} We evaluate the effects of parameters $s$, $k$ and $\delta$ in Fig.~\ref{fig:s,k}. From Fig.~\ref{fig:s,k}(a), we can see that multi-step mechanism is able to give about $1.5\%$, $1.9\%$ higher points to Rank-1 accuracy and mAP respectively. The performance is at its best when $s$ equals three. From Fig.~\ref{fig:s,k}(b) we can see that the parameter $k$ has a great impact on the performance. When we input the $5$-nearest neighbors features to the feature recovery transformer, FRT achieves the best result with Rank-1 accuracy of $70.7\%$ and mAP of $61.3\%$. From Fig.~\ref{fig:s,k}(c) we can find that when $\delta$ equals 0.2, it is able to give about $1.5\%$, $1.7\%$ higher points to Rank-1 accuracy and mAP respectively than $\delta$ equals 0. The reason is that the threshold $\delta$ can eliminate the effectiveness of occlusion during the training.

\textbf{Analysis of the Feature Recovery Transformer Time Consumption.} We evaluate the time consumption of the feature recovery transformer in Fig.~\ref{fig:time}. Fig.~\ref{fig:time} shows the time consumption on each image in inference. The result indicates that when $s$ equals three, the time consumption is about $0.12$ ms per image, which is nearly 1/40 of the Re-ranking costs. The results demonstrate that our feature recovery transformer is able to achieve better performance with less time consumption. 

\begin{table}[t]
\center
\caption{Evaluation of the feature recovery transformer on Occluded-Duke dataset. The result demonstrates that features are recovered after being processed by our feature recovery transformer.}
\setlength{\tabcolsep}{3.1mm}
\renewcommand\arraystretch{1.2}
\begin{tabular}{|c|c|c|c|c|}
\hline
\multirow{2}{*}{Feature} & \multicolumn{2}{c|}{\begin{tabular}[c]{@{}c@{}}Before\\ Feature Recovery\end{tabular}} & \multicolumn{2}{c|}{\begin{tabular}[c]{@{}c@{}}After\\ Feature Recovery\end{tabular}} \\ \cline{2-5}
 & Rank-1 & mAP & Rank-1 & mAP \\ \hline
Global & 55.6 & 46.1 & \textbf{69.3} & \textbf{59.1} \\
Head & 57.3 & 40.7 & \textbf{62.6} & \textbf{45.9} \\
Torso& 49.2 & 35.5 & \textbf{67.4} & \textbf{49.6} \\
Leg & 26.8 & 18.9 & \textbf{62.8} & \textbf{45.5} \\ \hline
Concat & 56.5 & 48.6 & \textbf{70.7} & \textbf{61.3} \\ \hline

\end{tabular}
\label{table:recovery}
\end{table}

\textbf{Evaluation of the Feature Recovery Transformer.} In Table~\ref{table:recovery}, we evaluate the Re-ID performance of each part feature and the final representation before and after being processed by feature recovery transformer on Occluded-Duke dataset. From the result we can find that before the feature recovery transformer, the leg part feature has the worst performance with Rank-1 accuracy of $26.8\%$ and mAP of $18.9\%$, indicating that the lower part of most of the images is occluded. After being processed by the feature recovery transformer, we can find that the Re-ID accuracy of the global feature, head feature, torso feature and leg feature are all improved. In particular, the Rank-1 accuracy and mAP of the leg feature are improved from $26.8\%$ to $62.8\%$ and $18.9\%$ to $45.5\%$ respectively, indicating that the occluded leg feature has been recovered by the feature recovery transformer. This experiment proves that the feature recovery transformer is able to recover occluded features and improve the Re-ID performance of both the global and local features.

\begin{figure}[t]
 \centering
 \includegraphics[width=\linewidth]{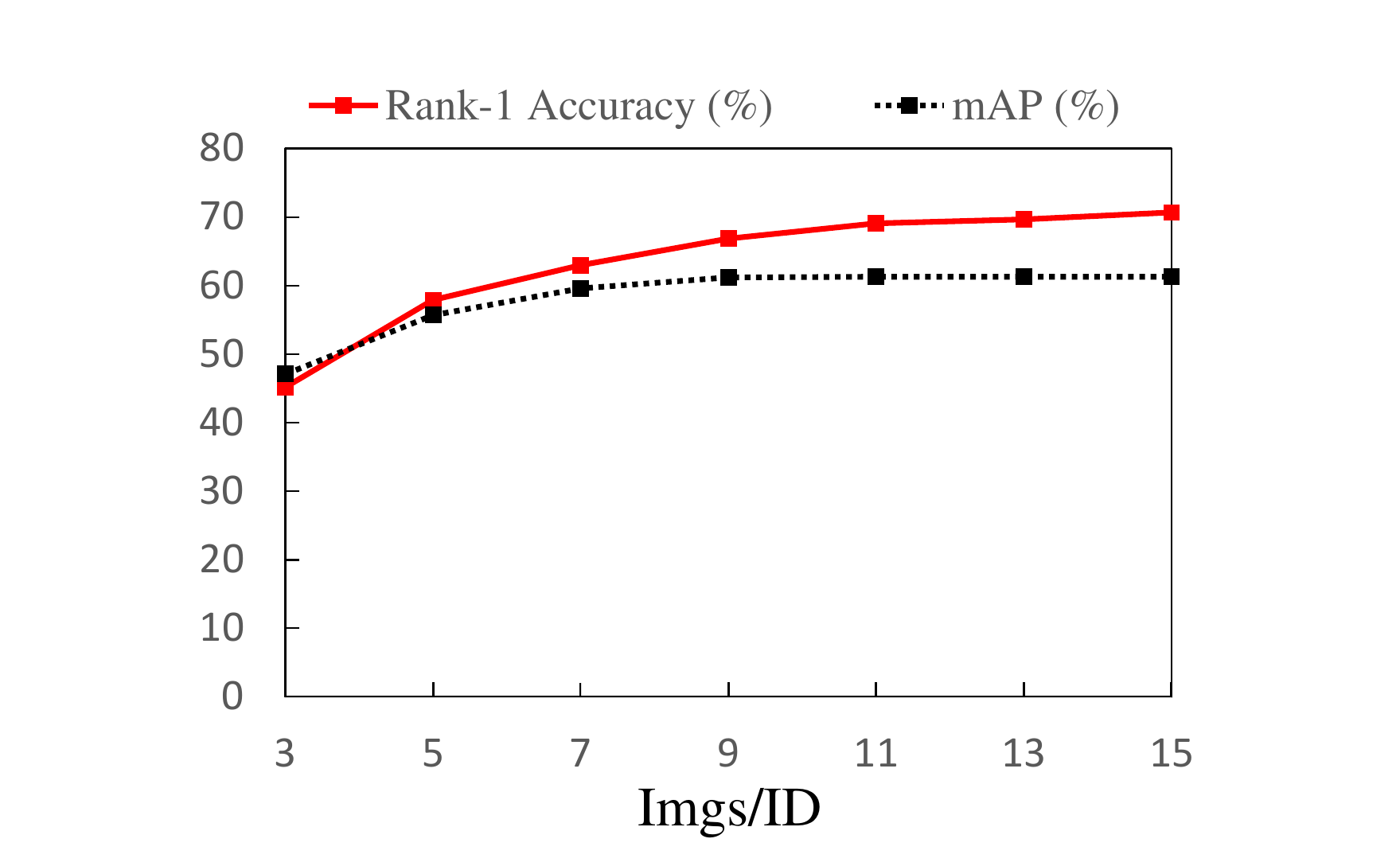}
 \caption{Evaluation of the effects of the gallery size. The gallery size is changed from 3 images per ID to 15 images per ID.}
 \label{gallery}
\end{figure}

\begin{figure}[h]
 \centering
 \includegraphics[width=\linewidth]{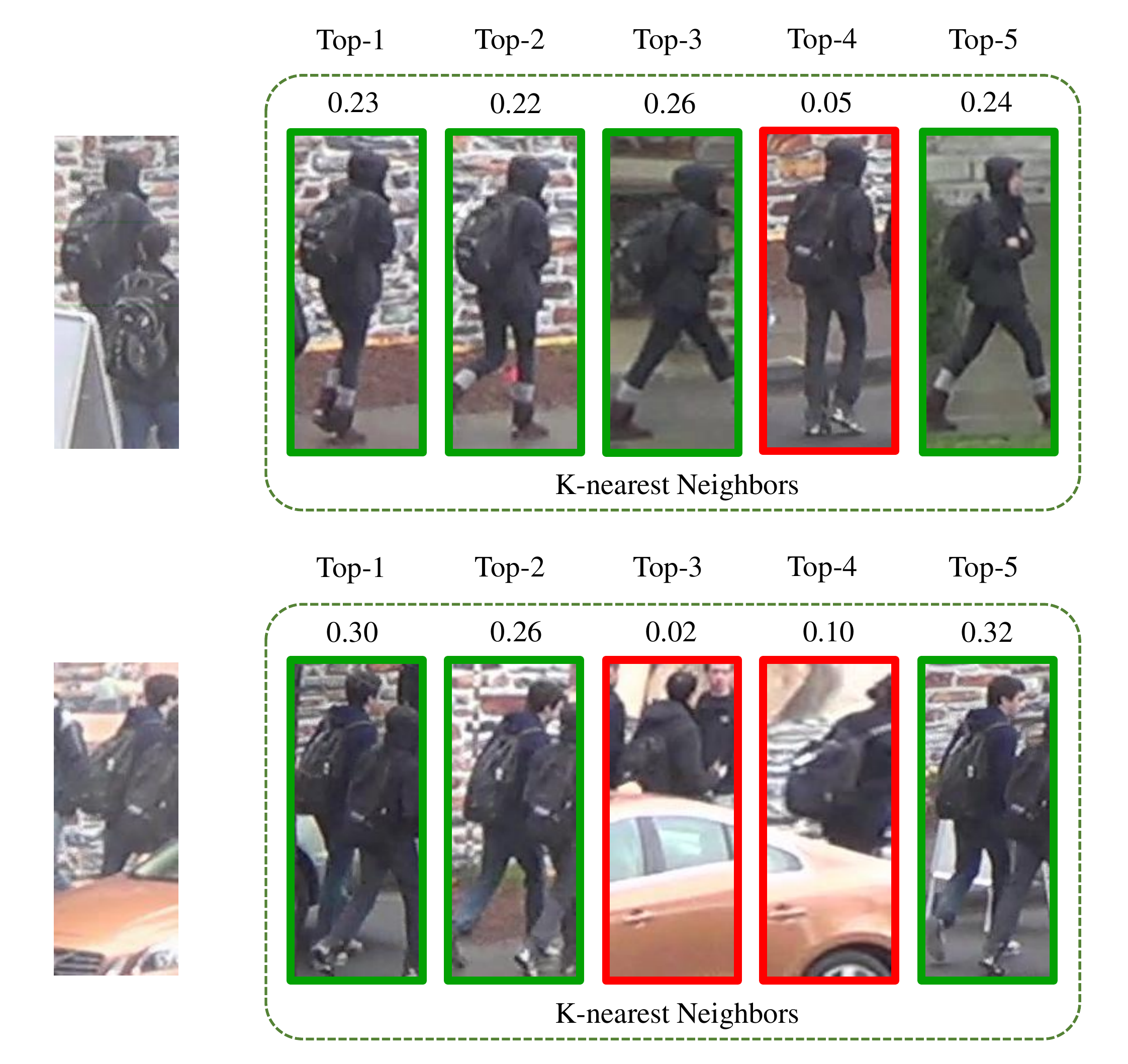}
 \caption{Visualization of the contributions of $k$-nearest neighbors when conducting the feature recovery transformer. We illustrate the top-5 nearest neighbors. Green and red rectangles indicate correct and error retrieval results,respectively. The numbers above the pictures indicate the contribution of corresponding neighbors to the feature recovery. The result shows that the feature recovery transformer is able to filter out the noise in the $k$-nearest neighbors and exploit valid information for feature recovery.}
 \label{visualtrans}
\end{figure}

\textbf{Visualization of the feature recovery transformer.} We visualize the recovery process of the feature recovery transformer in Fig.~\ref{visualtrans}. We illustrate the top-5 nearest neighbors. Green and red rectangles indicate correct and error retrieval results,respectively. The numbers above the pictures indicate the contribution of corresponding neighbors to the feature recovery. In the first example, the $k$-nearest neighbors contain an error retrieval with a similar appearance to the probe. The biggest difference between the correct and error retrieval is the shoes. During the process the feature recovery, we find that the $\mathcal{T}$ exploit little information from the error retrieval, and thus the recovered query feature is reliable for representing the probe identity. In the second example, the probe image is occluded by a car, which leads to the two error retrievals in the $k$-nearest neighbors. Fortunately, the $\mathcal{T}$ is able to distinguish the noise and exploit the message from the other 3 correct retrievals for the recovery of the occluded feature in the probe. The result shows that the feature recovery transformer is able to filter out the noisy information, even for some hard cases, in the $k$-nearest neighbors, and then exploit the valid message for feature recovery.

\textbf{Evaluation of the Effects of the Gallery Size.} Feature recovery transformer utilizes the pedestrian information in the gallery for features recovery. Therefor, we attempt to evaluate the effects of the gallery size on the feature recovery transformer. We change the gallery size from 3 images per identity to 15 images per identity, and the result is shown in Fig.~\ref{gallery}. From the result we can see that the performance of the FRT improves with the increase of the gallery size. Specifically, the Rank-1 accuracy and mAP are improved from $45.2\%$, $47.1\%$ to $70.7\%$, $61.3\%$ respectively when the gallery size increases from 3 images per ID to 15 images per ID. The reason is that when the gallery size increases, the feature recovery transformer is able to employ more pedestrian information in the gallery for features recovery.

\section{Conclusion}
In this paper, we propose a novel framework called Feature Recovery Transformer (FRT) for the occluded person re-identification. Firstly, we employ key-points to extract semantic features for alignment and calculate the visibility scores for them. Then, we consider the semantic features from same part within a pair of images as nodes to construct a directional graph. We set the edge based on the visibility score of starting nodes for promoting the propagation of information in the shared regions. In terms of the occluded feature recovery, we propose a feature recovery transformer to exploit the pedestrian information in the features of its $k$-nearest neighbors. Finally, the recovered query feature is utilized for retrieving. Extensive experiments on occluded, partial and holistic datasets demonstrate that our proposed framework is able to recover the occluded features and achieve the best Re-ID performance.

\section*{Acknowledgement}
We thank associate editor and anonymous reviewers for providing valuable suggestions to improve this paper.



\bibliographystyle{IEEEtran}
\bibliography{ref.bib}

\end{document}